\documentstyle[theapa,jair,11pt]{article}
 
\newtheorem{THEOREM}{Theorem}[section]
\newenvironment{theorem}{\begin{THEOREM} \hspace{-.85em} {\bf :} }%
                        {\end{THEOREM}}
\newtheorem{LEMMA}[THEOREM]{Lemma}
\newenvironment{lemma}{\begin{LEMMA} \hspace{-.85em} {\bf :} }%
                      {\end{LEMMA}}
\newtheorem{COROLLARY}[THEOREM]{Corollary}
\newenvironment{corollary}{\begin{COROLLARY} \hspace{-.85em} {\bf :} }%
                          {\end{COROLLARY}}
\newtheorem{PROPOSITION}[THEOREM]{Proposition}
\newenvironment{proposition}{\begin{PROPOSITION} \hspace{-.85em} {\bf :} }%
                            {\end{PROPOSITION}}
\newtheorem{DEFINITION}[THEOREM]{Definition}
\newenvironment{definition}{\begin{DEFINITION} \hspace{-.85em} {\bf :} \rm}%
                            {\end{DEFINITION}}
\newtheorem{CLAIM}[THEOREM]{Claim}
\newenvironment{claim}{\begin{CLAIM} \hspace{-.85em} {\bf :} \rm}%
                            {\end{CLAIM}}
\newtheorem{EXAMPLE}[THEOREM]{Example}
\newenvironment{example}{\begin{EXAMPLE} \hspace{-.85em} {\bf :} \rm}%
                            {\end{EXAMPLE}}
\newtheorem{REMARK}[THEOREM]{Remark}
\newenvironment{remark}{\begin{REMARK} \hspace{-.85em} {\bf :} \rm}%
                            {\end{REMARK}}
 
\newcommand{\thm}{\begin{theorem}}
\newcommand{\lem}{\begin{lemma}}
\newcommand{\pro}{\begin{proposition}}
\newcommand{\dfn}{\begin{definition}}
\newcommand{\rem}{\begin{remark}}
\newcommand{\xam}{\begin{example}}
\newcommand{\cor}{\begin{corollary}}

\newcommand{\ethm}{\end{theorem}}
\newcommand{\elem}{\end{lemma}}
\newcommand{\epro}{\end{proposition}}
\newcommand{\edfn}{\bbox\end{definition}}
\newcommand{\erem}{\bbox\end{remark}}
\newcommand{\exam}{\bbox\end{example}}
\newcommand{\ecor}{\end{corollary}}

\newcommand{\beqn}{\begin{equation}}
\newcommand{\eeqn}{\end{equation}}

\newcommand{\bbox}{\vrule height7pt width4pt depth1pt}

\newcommand{\clm}{\begin{claim}}
\newcommand{\eclm}{\end{claim}}

\newcommand{\inter}{\cap}

\newcommand{\IR}{\mbox{$I\!\!R$}}

\renewcommand{\phi}{\varphi}

\newcommand{\W}{{\cal W}}

\newcommand{\ol}{\setlength{\itemsep}{0pt}\begin{enumerate}}
\newcommand{\eol}{\end{enumerate}\setlength{\itemsep}{-\parsep}}
\newcommand{\ul}{\setlength{\itemsep}{0pt}\begin{itemize}}
\newcommand{\dl}{\setlength{\itemsep}{0pt}\begin{description}}
\newcommand{\edl}{\end{description}\setlength{\itemsep}{-\parsep}}
\newcommand{\eul}{\end{itemize}\setlength{\itemsep}{-\parsep}}

\newcommand{\commentout}[1]{}

\newcommand{\bi}{\begin{itemize}}
\newcommand{\ei}{\end{itemize}}
\newcommand{\be}{\begin{enumerate}}
\newcommand{\ee}{\end{enumerate}}

\jairheading{11}{1999}{429--435}{6/99}{12/99}
\ShortHeadings{Cox's Theorem Revisited}{Halpern}
\firstpageno{429}

\begin{document}

\newcommand{\Bel}{\mbox{Bel}}
\newcommand{\Belo}{\mbox{Bel}_0}
\newcommand{\Belod}{\mbox{Bel}_0}
\newtheorem{THEOREMA}{Theorem}
\newenvironment{theorema}{\begin{THEOREMA} \hspace{-.85em} {\bf :} }%
                        {\end{THEOREMA}}
\newcommand{\thma}{\begin{theorema}}
\newcommand{\ethma}{\end{theorema}}

\title{Cox's Theorem Revisited}
\author{\name Joseph Y.\ Halpern \email halpern@cs.cornell.edu\\
\addr   Cornell University, Computer Science Department\\
   Ithaca, NY 14853\\
   http://www.cs.cornell.edu/home/halpern
}
\technicaladdendum
\maketitle
\begin{abstract}
The assumptions needed to prove Cox's Theorem are discussed and
examined.  Various sets of assumptions under which a Cox-style theorem
can be proved are
provided, although all are rather strong and, arguably, not natural.
\end{abstract}

\bigskip

I recently wrote a paper \cite{Hal16}
casting doubt on how compelling a justification for probability
is provided by Cox's celebrated theorem \cite{Cox}.
I have received (what seems to me, at least) a surprising amount of
response to that article.  Here I attempt to clarify the degree to which
I think Cox's theorem can be salvaged and respond to a glaring
inaccuracy on my part pointed out by Snow \citeyear{Snow98}.
(Fortunately, it is
an inaccuracy that has no affect on either the correctness
or the interpretation of the results of my paper.)  I have tried to
write this note with enough detail so that it can be read independently
of my earlier paper,  but I encourage the reader to consult
the earlier paper as 
well as the two major sources it is based on \cite{Cox,Paris94},
for further details and discussion.

Here is the basic situation.  Cox's goal is to ``try to show that \ldots
it is possible to derive the rules of probability from two quite
primitive notions which are independent of the notion of ensemble and
which \ldots  appeal rather immediately to common sense'' \cite{Cox}.
To that end, he starts with a function $\Bel$ that
associates a real number with each pair $(U,V)$ of subsets of a domain
$W$ such that $U \ne
\emptyset$.  We write $\Bel(V|U)$ rather than $\Bel(U,V)$, since we
think of $\Bel(V|U)$ as the belief, credibility, or likelihood of $V$
given $U$.
Cox's Theorem as informally understood, states that if $\Bel$ satisfies
two very reasonable restrictions, then $\Bel$ must be isomorphic to a
probability measure.  The first one says that the belief in $V$
complement (denoted $\overline{V}$) given $U$ is a function of the
belief in $V$ given $U$; the second says that the belief in $V \inter
V'$ given $U$ is a function of the belief in $V'$ given $V \inter U$ and
the belief in $V$ given $U$.  Formally, we assume that there are
functions $S: \IR \rightarrow \IR$ and $F: \IR^2 \rightarrow \IR$ such
that
\begin{itemize}
\item[A1.] $\Bel(\overline{V}|U) = S(\Bel(V|U))$ if $U \ne \emptyset$,
for all $U, V \subseteq W$.
\item[A2.] $\Bel(V \inter V'|U)
= F(\Bel(V'| V \inter U), \Bel(V|U))$ if $V \inter U \ne \emptyset$, for
all $U, V, V' \subseteq W$.
\end{itemize}
If $\Bel$ is a probability measure, then we can take
$S(x) = 1-x$ and $F(x,y) = xy$.

Before going on, notice that Cox's result does not claim that $\Bel$ is
a probability measure, just that it is {\em isomorphic\/} to a
probability measure.  Formally, this means that there is
a continuous one-to-one onto
function $g: \IR \rightarrow \IR$ such
that $g \circ \Bel$ is a probability measure on $W$, and
\begin{equation}\label{eq7}
g(\Bel(V|U)) \times g(\Bel(U)) = g(\Bel(V \inter U)) \mbox{ if $U \ne
\emptyset$,}
\end{equation}
where $\Bel(U)$ is an abbreviation for $\Bel(U|W)$.

If we are willing to accept that belief is real valued (this is a
strong assumption since,
among other things, it commits us to the assumption that beliefs cannot
be
incomparable---for any two events $U$ and $V$, we must have either
$\Bel(U) \le \Bel(V)$ or $\Bel(V) \le \Bel(U)$),
then A1 and A2 are very reasonable.
If this were all it took to prove Cox's Theorem, then it indeed would be
a very compelling argument for the use of probability.

Unfortunately, it is well known
that A1 and A2 by themselves do not suffice to prove Cox's Theorem.
Dubois and Prade \citeyear{DP90} give an example of a
function $\Bel$, defined on a finite domain, that satisfies A1 and A2
with $F(x,y) = \min(x,y)$ and $S(x) = 1-x$ but
is not isomorphic to a probability measure.  Thus, if we are to
prove Cox's Theorem, we need to have additional assumptions.

It is hard to dig out of Cox's papers \citeyear{Cox,Cox1}
exactly what additional assumptions his proofs need.  I show in my paper
that the result is false under some quite strong assumptions (see
below).  My result also suggests that most of the other proofs given of
Cox-style theorems are at best incomplete (that is, they require
additional assumptions beyond those stated by the authors); see
my previous paper for discussion.  The goal of this note is to clarify what it
takes to prove a Cox-style theorem, by giving a number of hypotheses
under which the result can be proved.  All of the positive versions of
the theorem that I state can be proved in a straightforward way by
adapting the proof given by
Paris \citeyear{Paris94}.  (This is the one correct, rigorous proof
of the
result of which I am aware, with all the hypotheses stated clearly.)
Nevertheless, I believe it is worth identifying all these variants,
since they are philosophically quite different.

Paris \citeyear{Paris94} proves Cox's Theorem
under the following additional assumptions:
\begin{itemize}
\item[Par1.] The range of $\Bel$ is $[0,1]$.
\item[Par2.] $\Bel(\emptyset|U) = 0$ and $\Bel(U|U) = 1$ if $U \ne
\emptyset$.
\item[Par3.] The $S$ in A1 is decreasing.
\item[Par4.] The $F$ is A2 is strictly increasing (in each coordinate)
in $(0,1]^2$ and continuous.
\item[Par5.] For all $0 \le \alpha, \beta, \gamma \le 1$ and $\epsilon >
0$, there
are sets $U_1 \supseteq U_2 \supseteq U_3 \supseteq U_4$ such that $U_3
\ne \emptyset$, and each of
$|\Bel(U_4|U_3) - \alpha|$, $|\Bel(U_3|U_2)
- \beta|$, and $|\Bel(U_2|U_1) - \gamma|$
is less than $\epsilon$.
\end{itemize}

\thma\label{Paristhm} {\rm \cite{Paris94}} If Par1-5 hold, then $\Bel$ is
isomorphic to a probability measure. \ethma

There is nothing special about 0 and 1 in Par1 and Par2; all we need to
assume is that there is some interval $[e,E]$ with $e < E$ such that
$\Bel(V|U) \in
[e,E]$ for all $V, U \subseteq W$, $\Bel(\emptyset|U) = e$, and
$\Bel(U|U) = E$.  These assumptions certainly seem reasonable, provided
we accept that beliefs should be linearly ordered.
Nor is it hard
too hard to justify Par3 and Par4
(indeed, Cox justifies them in his original paper).  The problematic
assumption here is Par5 (called A4 in my earlier paper and Co5 by Paris
\citeyear{Paris94}).  Par5 can be thought of as a density requirement; among
other things, it says that for each fixed $V$, the set of values that
$\Bel(U|V)$ takes on is dense in $[0,1]$.  It follows that, in
particular, to satisfy Par5, $W$ must be infinite; Par5 cannot be
satisfied in finite domains.  While ``natural'' and
``reasonable'' are, of course, in the eye of the
beholder, it does not strike me as a natural or reasonable assumption in
any obvious sense of the words.  This is particularly true since many
domains of interest in AI (and other application areas) are finite; any
version of Cox's Theorem that uses Par5 is simply not applicable in
these domains.   Can we weaken Par5?

Cox does not require anything like Par5 in his paper.  He does require
at various times that $F$ be
twice differentiable, with a continuous second derivative, and
that $S$ be twice differentiable.%
\footnote{Cox never collects his assumptions in any one place, so it is
somewhat difficult to tell exactly what he thinks he needs for his
proof.  More on this later.}
While differentiability assumptions are perhaps not as compelling as
continuity assumptions, they do seem like reasonable technical
restrictions.  Unfortunately, the counterexample I give in my earlier
paper shows that these assumptions do {\em not\/} suffice to prove Cox's 
theorem.  What I show is the following.

\thma {\rm \cite{Hal16}}
There is a function $\Belo$, a finite domain $W$, and functions $S$ and
$F$ satisfying A1 and A2, respectively, such that
\begin{itemize}
\item $\Belo(V|U) \in [0,1]$ for $U \ne \emptyset$,
\item $S(x) = 1-x$ (so that $S$ is strictly decreasing and infinitely
differentiable),
\item $F$ is infinitely differentiable, nondecreasing
 in each argument in
$[0,1]^2$, and strictly increasing in each argument in
$(0,1]^2$.  Moreover, $F$ is commutative, $F(x,0) = F(0,x) = 0$, and
$F(x,1) = F(1,x) = x$.
\end{itemize}
However, $\Belo$ is not isomorphic to a probability measure.
\ethma

To understand what the makes counterexample tick and the role of Par5,
it is useful to review part of Cox's argument.
In the course of his proof, Cox shows that A2 forces
$F$ to be an associative function, that is, that
\begin{equation}\label{eq1}
F(x,F(y,z)) = F(F(x,y),z).
\end{equation}
Here is Cox's argument.

Suppose $U_1 \supseteq U_2 \supseteq U_3 \supseteq U_4$.
Let $x = \Bel(U_4|U_3)$, $y = \Bel(U_3|U_2)$, $z = \Bel(U_2|U_1)$, $u_1
= \Bel(U_4|U_2)$, $u_2= \Bel(U_3|U_1)$, and $u_3 = \Bel(U_4|U_1)$.  By
A2, we have that $u_1 = F(x,y)$, $u_2 = F(y,z)$, and $u_3 = F(x,u_2) =
F(u_1,z)$.  It follows that $F(x,F(y,z)) = F(F(x,y),z)$.

Note that this argument does {\em not\/} show that $F(x,F(y,z)) =
F(F(x,y),z)$ for all $x,y,z$.  It shows only that the equality  holds
for those $x,y,z$ for which there exist $U_1 \subseteq
U_2 \subseteq U_3 \subseteq U_4$ such that
$x = \Bel(U_1|U_2)$, $y = \Bel(U_2|U_3)$, and $z = \Bel(U_3|U_4)$.
Par5 guarantees that the set of such $x,y,z$ is dense in $[0,1]^3$.
Combined with the continuity of $F$ assumed in Par4, this tells us that
(\ref{eq1}) holds for all $x,y,z$.

I had claimed in my earlier paper that none of the authors who had proved
variants of Cox's Theorem, including Cox himself, Acz\'el,
and Reichenbach, seemed to be aware of the need to make (\ref{eq1})
hold for all $x,y,z$.%
\footnote{As I pointed out in in my earlier paper, Acz\'el recognized this
problem in later work \cite{AczelDaroczy}.}
I was wrong in including Cox in this list.  (This is the glaring
inaccuracy I referred to above.)
As Snow \citeyear{Snow98} points out, Cox actually does realize that
$F$ must satisfy (\ref{eq1}) for all $x,y,z$, and explicitly makes this
assumption at a certain point in his first paper \cite{Cox}, although he
does not make this assumption explicitly in his (more informal) later
paper \cite{Cox1}.

Unfortunately, although Cox escapes from my criticism by recognizing the
need to make this assumption, it does not make his theorem any less
palatable.  Indeed, if anything, it makes matters worse.  Associativity
is a rather strong assumption, as Cox himself shows. In fact, Cox
shows that if $F$ is associative and
has continuous second derivatives, then $F$ is isomorphic to
multiplication, that is, there exists a function $f$ and constant
$C$ such that $Cf[F(x,y)] = f(x) f(y)$.  Let me stress that the
conclusion that $F$  is isomorphic to multiplication just follows from
the fact that it is associative and has continuous second derivatives,
and has nothing to do with A2.  Of course, by the time we
are willing to assume that there is a function $F$ that is isomorphic to
multiplication that satisfies A2, then we are well on the way to showing
that $\Bel$ is isomorphic to a probability measure.  For future
reference, I remark that Paris shows (in his Lemma 3.7) that Par1, Par2,
Par4, and Par5 suffice to show that $F$ is isomorphic to multiplication
(and that we can take $C=1$).

In any case, suppose we are willing to strengthen
Par4 so
as to require $F$ to be associative as well as continuous and strictly
increasing.  Does this suffice to get rid of Par5 altogether?
Unfortunately, it does not seem to.

Later in his argument, Cox shows that $S$ must satisfy the following
two functional equations for all sets $U_1 \supseteq U_2 \supseteq U_3$:
\begin{equation}\label{eq2.5}
S[S(\Bel(U_2|U_1))] = \Bel(U_2|U_1)
\end{equation}
and
\begin{equation}\label{eq2}
\Bel(U_2|U_1) \, S(\Bel(U_3|U_1)/\Bel(U_2|U_1)) =
S[S(\Bel(U_2|U_1))/S(\Bel(U_3|U_1))] S(\Bel(U_3|U_1))
\end{equation}
This means that for all $x$ and $y> 0$ for which there exist sets
$U_1$, $U_2$, and $U_3$ such that $x = \Bel(U_3|U_1)$ and
$y = \Bel(U_2|U_1)$, we have
\begin{equation}\label{eq3}
S(S(y)) = y
\end{equation}
and
\begin{equation}\label{eq3.5}
 yS(x/y) = S(x)S[S(y)/S(x)].
\end{equation}
Cox actually wants these equations to hold for all $x$ and $y$.
Paris shows that this follows from Par1--5.  (Here is Paris's argument.
Using Par3, it can be shown that $S$ is continuous (see 
\cite[Lemma~3.8]{Paris94}).
This combined with Par5 easily gives us that (\ref{eq3}) holds for all
$y \in [0,1]$.  (\ref{eq3.5}) follows from Par5 and the fact that
$F$ must be isomorphic to multiplication; as I mentioned above, the
latter fact is shown by Paris to follow from Par1, Par2,
Par4, and Par5.)
Without Par5, we need to assume that
(\ref{eq3}) and (\ref{eq3.5}) both hold for all $x$ and $y$, and that
is what Cox does.%
\footnote{Actually, Cox starts with (\ref{eq2}) and derives the
more symmetric functional equation $yS[S(x)/y] = xS[S(y)/x]$, rather
than (\ref{eq3.5}). It is this latter functional
equation that he assumes holds for all $x$ and $y$.  If we replace $x$
by $S(x)$ everywhere and use (\ref{eq3}),
then we get (\ref{eq3.5}).}

In the proof given by Paris for Theorem~\ref{Paristhm}, the only
use made of Par5 is in
deriving the associativity of $F$ and the fact that $S$ satisfies
(\ref{eq3}) and (\ref{eq3.5}).  Thus, we immediately get the following
variant of Cox's Theorem.
\thma\label{Coxthm} If Par1-4 hold and, in addition, the $F$ in A2 is
associative
and the $S$ in A1 satisfies both (\ref{eq3}) and (\ref{eq3.5}) for all
$x,y \in [0,1]$, then $\Bel$ is isomorphic to a probability measure.
\ethma
I stress here that A1 and A2 place constraints only on how $F$ and $S$
act on the range of $\Bel$ (that is, on elements $x$ of the form
$\Bel(U)$ for some subset $U$ of $W$), while associativity, (\ref{eq3}),
and (\ref{eq3.5}) place constraints on the global behavior of $F$ and $S$,
that is, on how $F$ and $S$ act even on arguments not in the range of
$\Bel$.  The example I give in my earlier paper can be viewed as giving a
$\Bel$ for which it is possible to find $F$ and $S$ satisfying A1 and
A2, but there is no $F$ satisfying A2 which is associative on $[0,1]$.

We can get a variant even closer to what Cox \citeyear{Cox} shows by
replacing Par4  by the assumption that $F$ is twice differentiable.  Note
that we need to make some continuity, monotonicity, or differentiability
assumptions on $F$.  As I mentioned earlier, Dubois and Prade show there
is a $\Bel$ that is not isomorphic to a probability function for which
$S(x) = 1-x$ and $F(x,y) = \min(x,y)$.  The $\min$ function is
differentiable (and {\em a fortiori\/} continuous), but is not twice
differentiable, nor is it strictly increasing in each coordinate in
$(0,1]^2$ (although it is nondecreasing).

The advantage of replacing Par5 by the requirement that $F$ be
associative and that $S$ satisfy (\ref{eq3}) and (\ref{eq3.5}) is that
this variant of Cox's Theorem now applies even if $W$ is finite.
On the other hand, it is hard (at least for me) to view
(\ref{eq3.5}) as a ``natural'' requirement. While assumptions like
associativity for $F$ and idempotency for $S$ (i.e., (\ref{eq3})) are
certainly natural mathematical assumptions, the only justification for
requiring them on all of $[0,1]$ seems to be that they provably
follow from the other assumptions for certain tuples in the range of
$\Bel$.  Is this reasonable or compelling?  Of course, that is up to the
reader to judge.  In any case, these are assumptions that needed to be
highlighted by anyone using Cox's Theorem as a justification for
probability, rather than being swept under the carpet.
The requirement that $S$ must satisfy (\ref{eq3.5}) is not even
mentioned by Snow \citeyear{Snow98}, let alone discussed.  Snow is not
alone; it does not seem to be mentioned in any other discussion of Cox's
results either (other than by Paris).   Of course, we can avoid
mentioning (\ref{eq3}) and (\ref{eq3.5}) by just requiring that
$S(x) = 1-x$ (as Cox \citeyear{Cox1} does).  However, this makes the
result less compelling.

A number of other variants of Cox's Theorem which are correct are
discussed in \cite[Section~5]{Hal16}.
Let me conclude by formalizing two of them that apply to finite domains,
but use Par5 (or slight variants of it), rather than assuming that $F$
must
be associative and that $S$ must satisfy (\ref{eq3}) and (\ref{eq3.5})
for all pairs $x, y \in [0,1]$.

The first essentially assumes that we can extend any finite domain to an
infinite domain by adding a sufficiently many ``irrelevant''
propositions, such as the tosses of  fair coin.  As I observed in
my earlier paper, this type of extendability argument is fairly standard.
For example, it is made by Savage \citeyear{Savage} in the course of
justifying one of his axioms for preference.  Snow \citeyear{Snow98}
essentially uses it as well.  Formally, this gives us the following
variant of Cox's Theorem, whose proof is a trivial variant of that of
Theorem~\ref{Paristhm}.

\thma\label{variant1}  Given a function $\Bel$ on a domain $W$,
suppose there exists a domain $W^+ \supseteq W$ and a function $\Bel^+$
extending $\Bel$ defined on all subsets of $W^+$ such that A1 and A2 hold
for $\Bel^+$ and all subsets $U, V, V'$ of $W^+$ and Par1-5 hold for
$\Bel^+$.  Then $\Bel^+$ (and hence $\Bel$) is isomorphic to a probability
measure.
\ethma
The problem with this approach is that it requires us to extend $\Bel$
to events we were never interested in considering in the first place,
and to do so in a way that is guaranteed to continue to satisfy Par1-5.

The second variant assumes that $\Bel$ is defined not just on one domain
$W$, but on all domains (or at least, a large family of domains); the
functions $F$ and $S$ then have to be uniform across all domains.
More precisely, we would get the following.

\thma\label{variant2} Suppose we have a function $\Bel$ defined on all
domains
$W$ in some set $\W$ of domains, there exist functions $F$ and $S$ such
that $F$ and $S$ satisfy A1 and A2 for all the domains $W \in \W$,
Par1--4 hold for $F$ and $S$, and
the following variant of Par5 holds:
\begin{itemize}
\item[Par5$'$.] for all $0 \le \alpha, \beta, \gamma \le 1$ and
$\epsilon > 0$, there exists $W \in \W$ and sets $U_1, U_2, U_3, U_4 \subseteq W$
such that
$U_1 \supseteq U_2 \supseteq U_3 \supseteq U_4$, $U_3
\ne \emptyset$, and each of
$|\Bel(U_4|U_3) - \alpha|$, $|\Bel(U_3|U_2)
- \beta|$, and $|\Bel(U_2|U_1) - \gamma|$
is less than $\epsilon$.
\end{itemize}
Then $\Bel$ is {\em uniformly isomorphic\/} to a probability measure, in
that there exists a function
$g: \IR \rightarrow \IR$ such that for all $W \in \W$,
we have that $g \circ \Bel$ is a probability measure on each $W$ and for
all
$U,\, V \subseteq W$, we have
$$g(\Bel(V|U)) \times g(\Bel(U)) = g(\Bel(V \inter U)) \mbox{ if $U \ne
\emptyset$.}$$
\ethma

The advantage of this formulation is that $\W$ can consist of only
finite domains; we never have to venture into the infinite (although
then $\W$ would have to include infinitely  many finite domains).
This conception of one function $\Bel$ defined uniformly over a family
of domains seems consistent with the philosophy of both Cox and Jaynes
(see, in particular, \cite{Jaynes97}).

While the hypotheses of Theorems~\ref{variant1} and~\ref{variant2} may
seem more reasonable than some others (at least, to some readers!), note
that they still both essentially require Par5 and, like all the other
variants of Cox's Theorem that I am aware of, disallow a
notion of belief that has only finitely many gradations.
One can justify a notion of belief that takes on all values in $[0,1]$
by continuity considerations (again, assuming that one accepts a
linearly-ordered notion of belief), but it is still a nontrivial
requirement.%
\footnote{Snow \citeyear{Snow98} quotes the conference version of
\cite{Hal16} (which appeared in AAAI '96, pp.~1313--1319) as saying `Cox's
Theorem ``disallows a notion of belief that takes on only finitely many
or countably many gradations'',' but what I say disallows a notion of
belief is not Cox's Theorem, but the viewpoint that assumed that $\Bel$
varies continuously from 0 to 1.
In fact, Co5 is compatible with a notion of belief that takes on countably many
(although not finitely many) values.
(Essentially the same sentence appears in the journal version of the
paper, where it does refer to Cox's Theorem, but
without the phrase ``or countably''.)}

I will stop at this point and leave it to the reader to form his or her
own beliefs.

\section*{Acknowledgments}
I'd like to thank Paul Snow for some useful email exchanges on this
topic (and for pointing out that Cox had in fact realized the need to
assume that $F$ was associative for all $(x,y,z)$).
This work was supported in part by the NSF, under grant
IRI-96-25901.

\bibliographystyle{theapa}
\bibliography{z,joe}
\end{document}